\documentclass[final]{cvpr}

\usepackage{times}
\usepackage{epsfig}
\usepackage{graphicx}
\usepackage{amsmath}
\usepackage{amssymb}
\usepackage{subfigure}

\usepackage[pagebackref=true,breaklinks=true,colorlinks,bookmarks=false]{hyperref}

\def\cvprPaperID{8} 
\def\confYear{ICCV Workshop 2021}
\begin{document}

\title{Multiple GAN Inversion for Exemplar-based Image-to-Image Translation}

\author{
	Taewon Kang \\
	Department of Computer Science and Engineering \\ Korea University, Seoul, Korea\\ 
	\texttt{itschool@itsc.kr} \\
} 

\maketitle

\begin{abstract}
   Existing state-of-the-art techniques in exemplar-based image-to-image translation hold several critical concerns. Existing methods related to exemplar-based image-to-image translation are impossible to translate on an image tuple input (source, target) that is not aligned. Additionally, we can confirm that the existing method exhibits limited generalization ability to unseen images. In order to overcome this limitation, we propose Multiple GAN Inversion for Exemplar-based Image-to-Image Translation. Our novel Multiple GAN Inversion avoids human intervention by using a self-deciding algorithm to choose the number of layers using Fréchet Inception Distance(FID), which selects more plausible image reconstruction results among multiple hypotheses without any training or supervision. Experimental results have in fact, shown the advantage of the proposed method compared to existing state-of-the-art exemplar-based image-to-image translation methods.
\end{abstract}

\section{Introduction}
Many computer vision problems can be viewed as an Image-to-Image Translation problem, such as style transfer~\cite{gatys2016image,huang2017arbitrary}, super-resolution~\cite{dong2015image,kim2016accurate}, image inpainting~\cite{pathak2016context,iizuka2017globally}, colorization~\cite{zhang2016colorful, zhang2017real} and so on. Essentially, image-to-image translation is a method to map an input image from one domain to a comparable image in a different domain. 

\begin{figure}[t]
	\begin{center}
	\renewcommand{\thesubfigure}{}
	\subfigure[Source]
	{\includegraphics[width=0.33\linewidth]{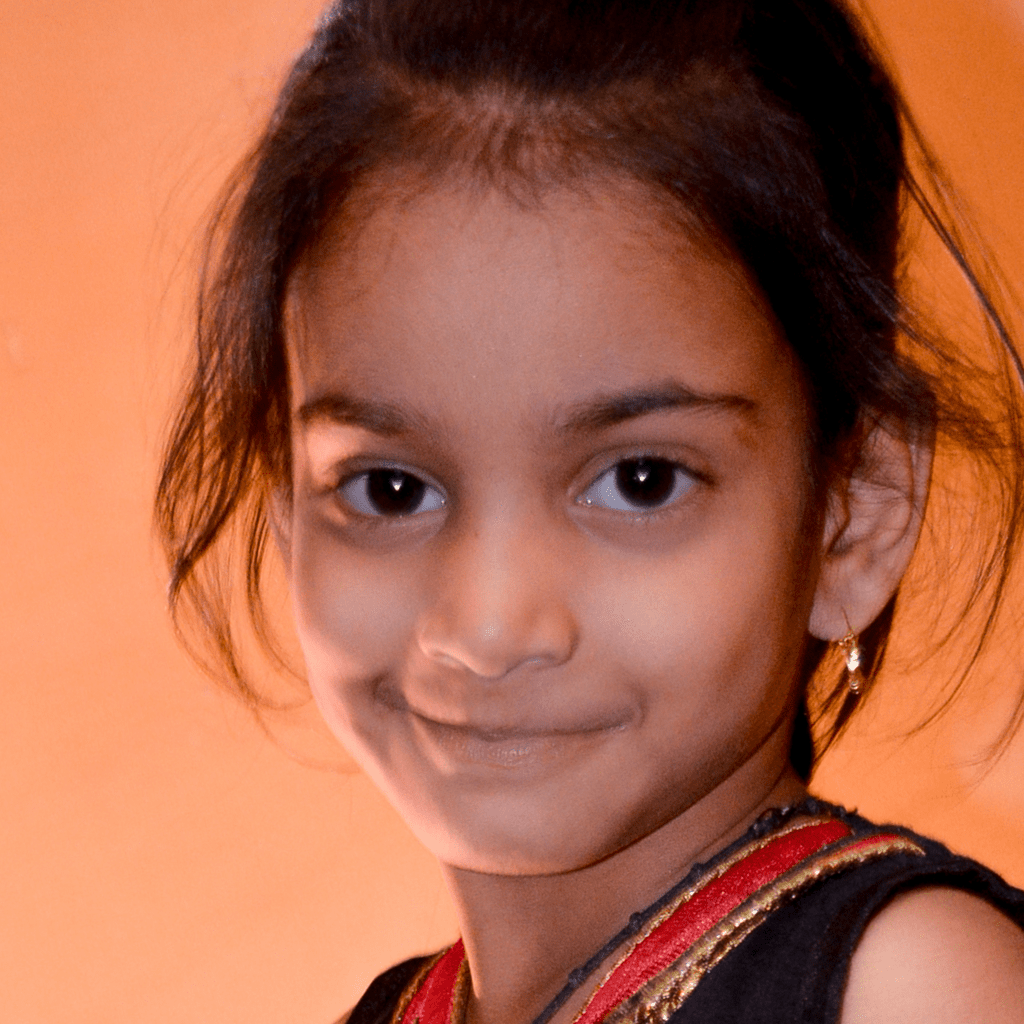}}\hfill
	\subfigure[Warped Image]
	{\includegraphics[width=0.33\linewidth]{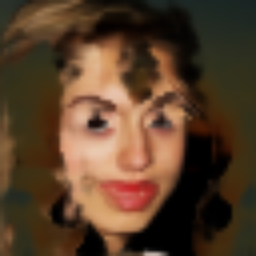}}\hfill	
	\subfigure[CoCosNet-final]
	{\includegraphics[width=0.33\linewidth]{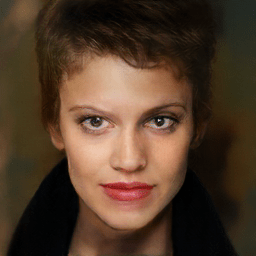}}\hfill \\
	\vspace{-5pt}
	\subfigure[Target]
	{\includegraphics[width=0.33\linewidth]{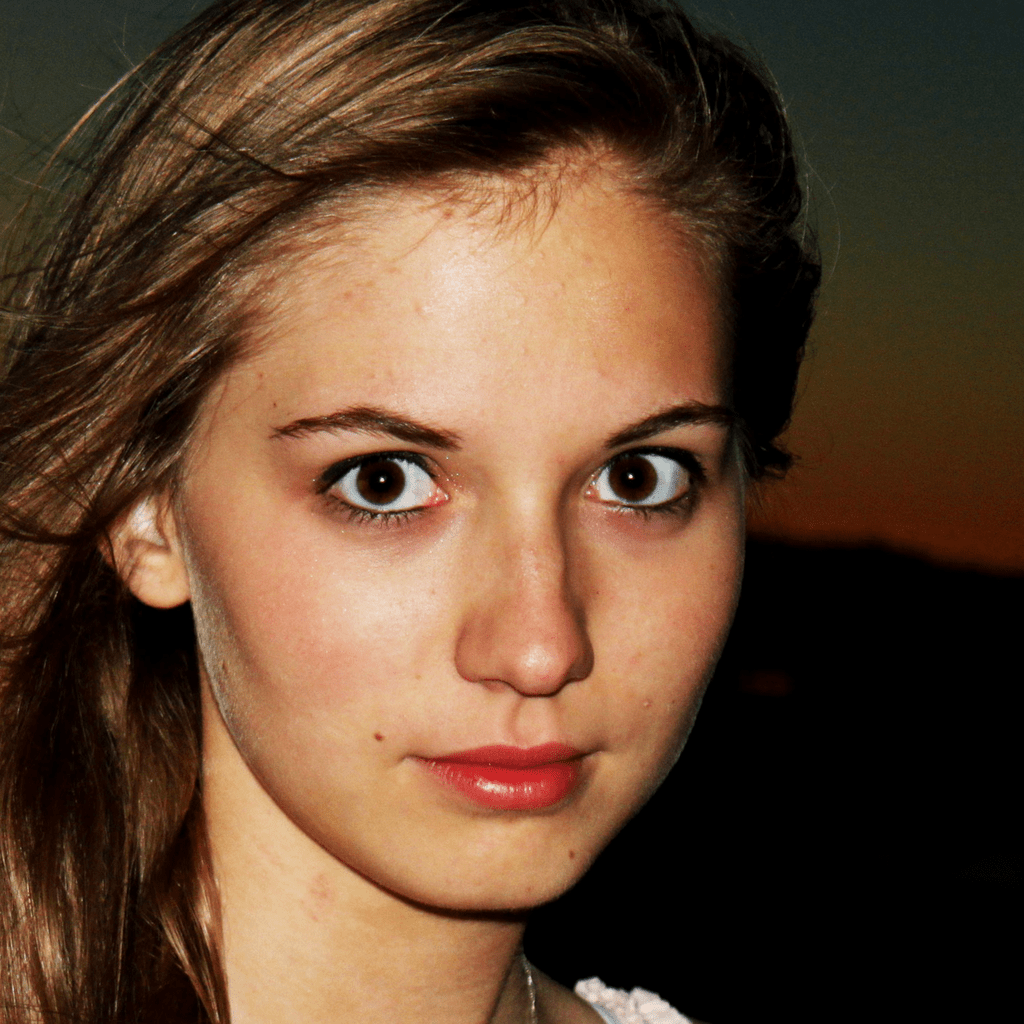}}\hfill
	\subfigure[OEFT-final]
	{\includegraphics[width=0.33\linewidth]{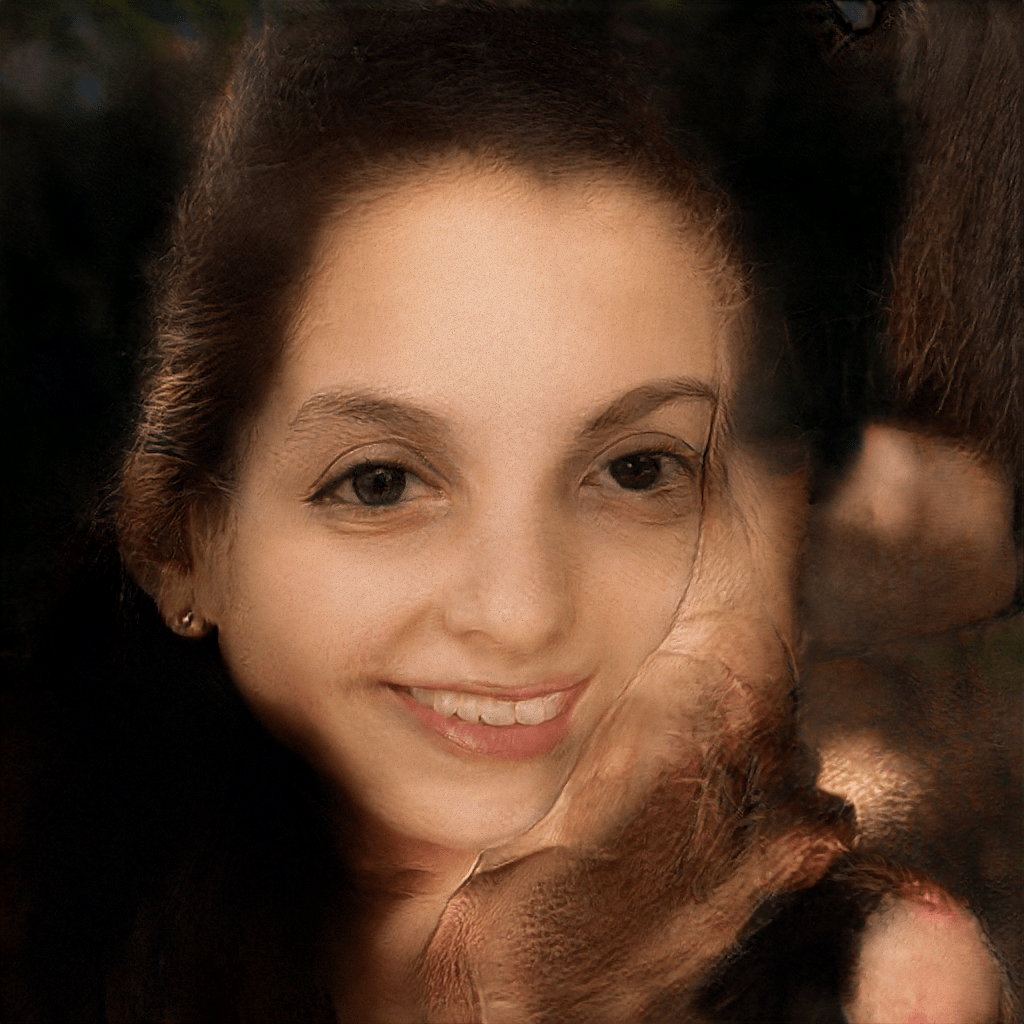}}\hfill
	\subfigure[\textbf{Ours-final}]
	{\includegraphics[width=0.33\linewidth]{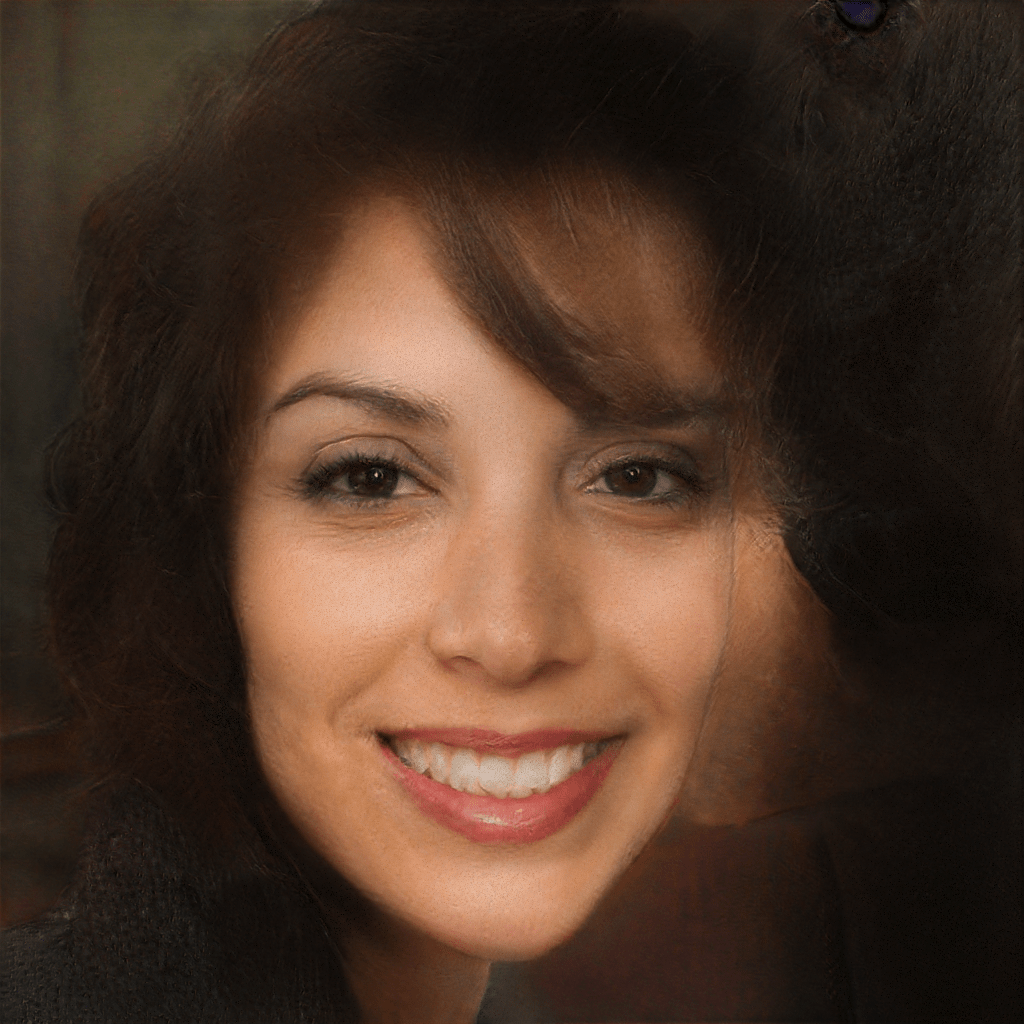}}\hfill\\
    \end{center}
    \vspace{-10pt}
	\caption{\textbf{Examples of exemplar-based image-to-image translation results using our network:} Existing methods, such as CoCosNet~\cite{zhang2020cross} and OEFT~\cite{kang2020online}, one of the state-of-the-art exemplar-based I2I shows a limited generalization power to unseen input pairs, despite its intensive training on large-scale dataset. Unlike these techniques, our framework proposes Multiple GAN Inversion(MGI) for exemplar-based image-to-image translation, having better generalization ability without training on specific image-to-image translation datasets.}
	\label{mgi_generation}
\end{figure}

Recently, there are several ways to solve exemplar-based image-to-image translation using the state-of-the-art technique~\cite{huang2018multimodal, ma2018exemplar, wang2019example, zhang2020cross}. Example-based I2I is a conditional image transformation that reconstructs the image of the source according to the guidance of the example on the target. For instance, SPADE~\cite{park2019semantic} shows impressive results when using Spatially-Adaptive Normalization as it improves the input semantic layout while performing affine transformations in the normalization layer. This model allows the user to select an external style image to control the ``global appearance'' of the output image while demonstrating the user control demos over both semantic and style when compositing images. 

CoCosNet~\cite{zhang2020cross} shows the most advanced results in exemplar-based image-to-image translation. However, despite the success of CoCosNet in this endeavour, it exhibits several critical problems. Firstly, training weights are required, leading to expensive computation costs. Secondly, such training weights are mandatory for each and every task (ex. mask-to-face, edge-to-face, pose-to-edge) alongside large-scaled training sets. Additionally, the SPADE block inside the CoCosNet network requires labeled masks for the training set, leading to restrictions in the practical use of the model. Finally, CoCosNet fails to match instances within a pair of images and causes the generated image to be overfitted to the mask image.

Recently, there have been many attempts to reverse the generation process by re-mapping the image space into a latent space, widely known as GAN inversion. For example, on the ``steerability'' of generative adversarial networks showed impressive results by exploring latent space. This paper shows how to achieve transformations in the output space by moving in latent space. Image2StyleGAN shows the StyleGAN inversion model (using pre-trained StyleGAN model) using efficient embedding techniques to map input image to the extended latent space W+. mGANprior shows the GAN Inversion technique using multiple latent codes and adaptive channel importance. Multiple latent codes allow the generation of multiple feature maps at some intermediate layer of the generator, then composes them with adaptive channel importance to recover the input image. However, mGANprior fails to generate the best results using the same layer and parameter for each data set.

In this paper, we explore an alternative solution, called Multiple GAN Inversion for Exemplar-based Image-to-Image Translation, to overcome aforementioned limitations for exemplar-based image-to-image translation. Multiple GAN Inversion(MGI) refines the translation hypothesis that self-decides to decipher the optimal translation result using warped image in exemplar-based I2I. Applying Super Resolution with GAN Inversion, we generated high-quality translation results. Check out the  \href{http://itsc.kr/2021/06/24/mgi2021/}{project page for more information.}

\begin{figure*}[t]
\begin{center}
\includegraphics[width=1\linewidth]{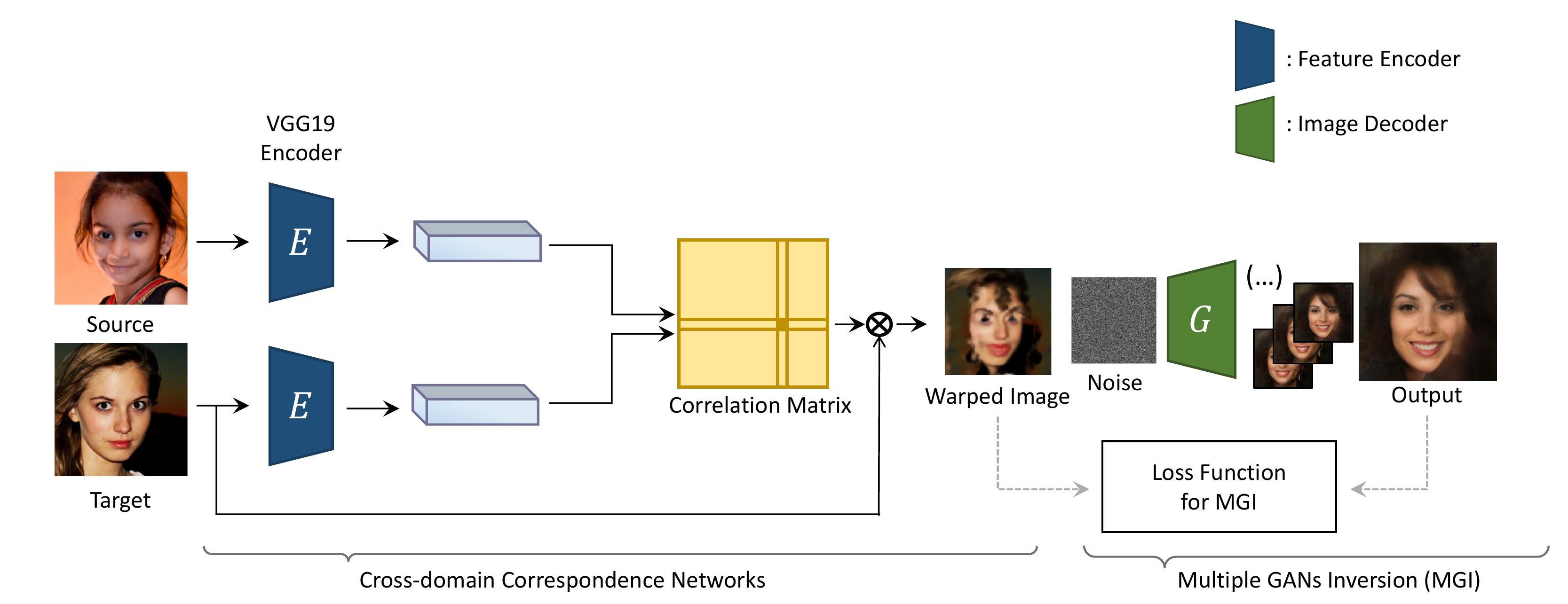}
\end{center}\vspace{-10pt}
  \caption{\textbf{Overall Architecture.} 
  Our architecture consists of two modules, namely cross-domain correspondence networks and multiple gan inversion. At the first, given the source and target image, the cross-domain correspondence submodule adapts them into same domain, where dense correspondence can be established. At the second, we refine the warped image to more natural and plausible one through the multiple GANs inversion.} \vspace{-10pt}
\label{overall}
\end{figure*}

\section{Related Works}
\textbf{Image-to-Image Translation} Image-to-Image translation aims to learn the mapping between two domains, which is based on the Generative Adversarial Networks(GANs)~\cite{goodfellow2014generative}. pix2pix~\cite{isola2017image} as the representative Supervised Image-to-Image Translation model. The pix2pix framework uses a conditional generative adversarial network to learn the mapping from input to output images. The discriminator, D, learns to classify between fake (synthesized by the generator), real edge and photo tupled images, while the generator, G, learns to fool the discriminator. Unlike an unconditional GAN, both the generator and discriminator observe the input edge map. CycleGAN~\cite{zhu2017unpaired} is the first Unsupervised Image-to-Image Translation algorithm which proposes the cycle consistency. The translation networks for both directions are trained together and they provide supervision signals for one another. 

UNIT~\cite{liu2017unsupervised} is one of the Unimodal Image-to-Image Translation algorithms. Unlike CycleGAN, UNIT proposes a different assumption called ‘shared latent space’. This assumes that the latent space can be shared by both domains, for instance, that two analogous images can map to the same latent code. MUNIT~\cite{huang2018multimodal} is a representative Multimodal Image-to-Image Translation model. MUNIT assumes that the image representation can be decomposed into a content code that is domain-invariant and a style code that captures domain-specific properties. For example, MUNIT can translate edge2shoes dataset containing example images (like edges of shoes) and shoes images, translating real shoes images to example edge images.

\textbf{Exemplar-based Image-to-Image Translation} Exemplar-based image-to-image translation is a kind of image-to-image translation method, which makes use of a structure image (the input) and style image (the exemplar) in order to generate an output. The point is that the output should contain both structure from input and style from exemplar which is ‘balanced’, and thus should be ‘natural’. 

Moderately balanced output has been an issue in exemplar image-to-image translation as it needs well-extracted features for both structure and style. Recent approaches require a paired image dataset or labeled image~\cite{zhang2020cross,park2019semantic,zhu2020sean} to catch semantic attributes more accurately. Some other approaches generate pretext in order to create a ground-truth-like image.

SPADE,~\cite{park2019semantic} which replaces normalization into a mixture of batch normalization and structure-based conditional form into a new form for taking more spatial information, fails to represent style. SEAN~\cite{zhu2020sean} points out two shortcomings of SPADE; firstly, that it has only one style code to control the entire style of an image and secondly that it inserts style information only in the beginning. To maintain more style information, SEAN introduces one style code per region, directing that style information to multiple locations. However, both SPADE and SEAN require labeled mask semantic information; in layman terms, this means they need a large scale of a labeled dataset. Our approach does not hold any requirements for labeled data, or even datasets. What we need is simply a couple of images as inputs and an exemplar.

Swapping Autoencoder~\cite{park2020swapping} has tried to resolve the balance problem by training a swapping autoencoder with two independent components: structure code and texture code. This model is innovative as it does not need any semantic information prior. However, Swapping Autoencoder fails to extract local style features and thus, we cannot say it has a moderately balanced output. Contrastingly, our work applies correlation matrix and GAN Inversion to extract local style features unsupervised.

CoCosNet~\cite{zhang2020cross} shows outperforming results in exemplar-based I2I, yet CoCosNet still captures several critical problems. Firstly, weight training is required so this leads to expensive computation costs. Secondly, such weighty training is mandatory for each and every type of task (ex. mask-to-face, edge-to-face, pose-to-edge) whilst also requiring large-scaled training sets. And finally, SPADE blocks inside the CoCosNet network require labeled masks for the training set, leading to restrictions in the practical use of the model. 

To address this problem, OEFT~\cite{kang2020online} OEFT(Online Exemplar Fine-Tuning) utilizes pre-trained off-the-shelf networks, and simply fine-tunes online for a single pair of input images, which does not require the off-line training phase. Correspondence fine-tuning module establishes accurate correspondence fields by solely considering the internal matching statistics. However, OEFT still displays several critical concerns regarding computation time and unseen input image translation generalizability. In stark contrast, our model shows the high generalization ability to any unseen input images and depicts less computation time in comparison to other approaches.

\textbf{GAN Inversion}
The goal of GAN inversion is to return a given image to a latent code using pre-trained GAN (PGGAN, StyleGAN, etc) model. 

Image2StyleGAN~\cite{abdal2019image2stylegan} is an efficient embedding technique that allows to map input image to the extended latent space W+ of a pre-trained StyleGAN model. This assumes multiple inquiries regarding insight on the structure of the StyleGAN latent space. Moreover, this algorithm provides style transfer, face embedding, etc.

PULSE~\cite{menon2020pulse} is a novel super-resolution GAN Inversion technique, which creates realistic high-resolution images that downscales to correct low resolution input. This algorithm proposes a new framework for single image super-resolution by using downscaling loss and latent space exploration.

Collaborative Learning for Faster StyleGAN Embedding~\cite{guan2020collaborative} uses the same design with Image2StyleGAN, but this model exhibits few differences, such as the initialization method, and the LPIPS loss method. Image2StyleGAN runtime is 420.00, but this paper runtime is 0.71, a 0.016\% faster runtime compared to Image2StyleGAN. 

mGANprior~\cite{gu2020image} provides multiple latent codes and adaptive channel importance in GAN Inversion through the assumption that an effective GAN Inversion method and mGANprior can be used in many applications; for example, super resolution, colorization, inpainting, semantic manipulation, etc. However, mGANprior fails to prevent optimized hyperparameters in different images. 

In our work, we propose Multiple GAN Inversion using mGANprior as it allows us to find the best hyperparameter in the GAN Inversion model. This avoids human intervention through the use of a self-deciding algorithm which chooses the number of layers using FID.

\section{Methods}
We want to learn the translation from the source domain($x_\mathcal{S} \in \mathcal{S}$) to the target domain($x_\mathcal{T} \in \mathcal{T}$). The generated output($r_{\mathcal{T} \rightarrow \mathcal{S}}$) is desired to conform to the content($x_\mathcal{S}$) while resembling the style from semantically similar parts in  style($x_\mathcal{T}$). To address this problem, we first establish a correspondence between $x_\mathcal{S}$ and $x_\mathcal{T}$ in different domains, and then warp the exemplar image accordingly to match the meaning with $x_\mathcal{S}$. After that, we refine the warped image, which is generated by cross-domain correspondence networks, in order to generate a more natural and plausible image through the proposed Multiple GAN Inversion, done using a self-deciding algorithm to choose the hyper-parameters for GANs inversion. Figure~\ref{overall} shows our pipeline of exemplar-based image-to-image translation.

\subsection{Cross-domain Correspondence Networks}
We first encode the source image($x_\mathcal{S}$) and the target image($x_\mathcal{T}$) using feature extractors and pre-trained VGG-19 feature extractors, to get feature($x_\mathcal{S}$, $x_\mathcal{T}$) as shown below:

\begin{figure}[t]
\begin{center}
\includegraphics[width=1\linewidth]{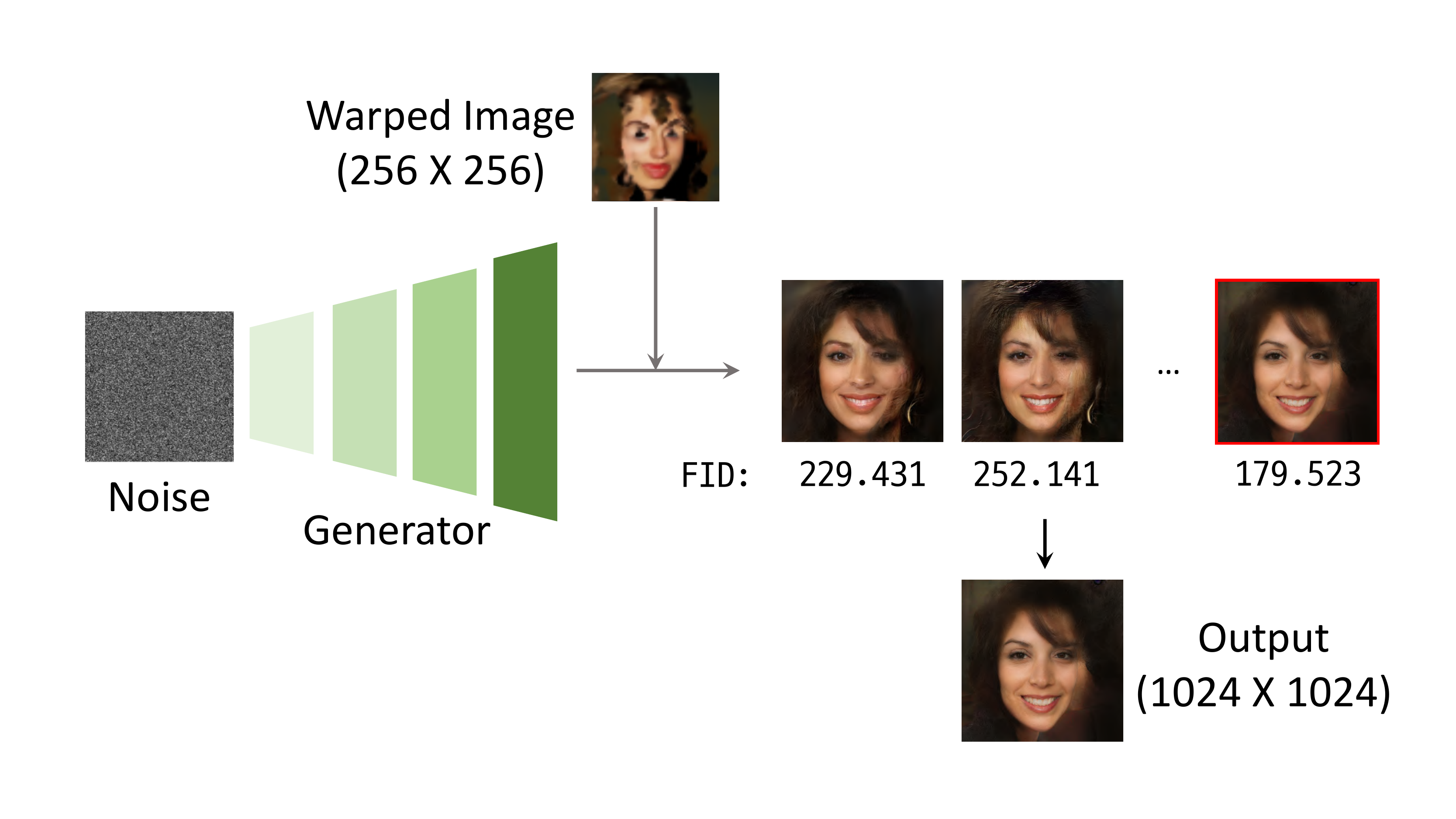}
\end{center}
\vspace{-18pt}
  \caption{\textbf{Illustration of Multiple GAN Inversion.} Our multiple GAN inversion network generates high-resolution images from a warped image, and this model can generate high-quality results from a low-resolution warped image. Multiple GAN Inversion also includes an innovative self-deciding algorithm in choosing the hyperparameters using FID that avoids human intervention.}
\label{mgan_architecture}
\end{figure}

\begin{align}
    f_\mathcal{S} &= F_\mathcal{S}(x_\mathcal{S};\mathbf{W}_\mathcal{S}) \in \mathbb{R}^{H\times W\times C}, \\
    f_\mathcal{T} &= F_\mathcal{T}(x_\mathcal{T};\mathbf{W}_\mathcal{T}) \in \mathbb{R}^{H\times W\times C}, 
\end{align}

$F(x;\mathbf{W})$ denotes a feed-forward process of input $x$ through a deep network of parameters $\mathbf{W}$. $H$, $W$ and $C$ denotes spatial size of the feature and channels. $\mathbf{W}_\mathcal{S}$ and $\mathbf{W}_\mathcal{T}$ are network parameters for encoding source and target features, respectively. \\

\textbf{Local Feature Matching} Ideally, the source and the warped image should structurally match. To address this problem, we propose a correlation matrix to warp the features of $x$ and $y$. Specifically, we need to calculate a global correlation matrix M $\in \mathbb{R}^{HW \times HW}$, in which each factor is a pairwise feature correlation.

\begin{equation}
    \mathcal{M}(u,v) = \frac{\hat{f}_\mathcal{S}(u)^T \hat{f}_\mathcal{T}(v)}{|\hat{f}_\mathcal{S}(u)|\cdot|\hat{f}_\mathcal{T}(v)|},
\end{equation}

$\hat{f}_\mathcal{S}(u)$ and $\hat{f}_\mathcal{T}(v)$ are the channel-wise centralized features of $f_\mathcal{S}$ and $f_\mathcal{T}$ at the position of u and v. $M(u, v)$ can be summarized as follows:

\begin{align}
    \hat{f}_\mathcal{S}(u) &= f_\mathcal{S}(u) - \text{mean}(f_\mathcal{S}), \\
    \hat{f}_\mathcal{T}(v) &= f_\mathcal{T}(v) - \text{mean}(f_\mathcal{T}), 
\end{align}

After that, we need to get the warped image into the correlation matrix, so we need to warp $x_\mathcal{T}$ according to $\mathcal{M}$ and obtain the warped exemplar $r_{\mathcal{T} \rightarrow \mathcal{S}} \in \mathbb{R}^{HW}$. More specifically, we need to choose the most relevant pixel from $x_\mathcal{T}$ and calculate the weighted average. This can be summarized as follows:



\begin{equation}
    r_{\mathcal{T} \rightarrow \mathcal{S}} (u) = \sum_{v}^{} \underset{v}{\mathrm{softmax}} (\alpha M(u,v)) \cdot x_\mathcal{T} (v) 
\end{equation}

\subsection{Multiple GANs Inversion}
Even though the previously mentioned technique produces a translation image, the output will inherently have a limited resolution(256$\times$256) due to memory limitations and some artifacts. To overcome these, CoCosNet~\cite{zhang2020cross} attempts to train additional translation networks using the SPADE block. However, it requires additional resources. We propose a method to utilize the GAN inversion~\cite{gu2020image}, in which we only optimize the latent code that is likely to generate the plausible image guided from the warped image $r_{\mathcal{T} \rightarrow \mathcal{S}}$ with the pre-trained, fixed generation network of parameters $\mathbf{W}_{I}$. Even though any GAN inversion technique can be used, in this paper, we specifically use a recent GAN Inversion technique, named mGANprior~\cite{gu2020image}. 

To further improve the performance from what was observed within this paper, we present a self-deciding algorithm when choosing the hyperparameters of GAN inversion, based on Fréchet Inception Distance (FID)~\cite{heusel2017gans}, which is a no-reference quality measure for image generation, so this algorithm does not need any supervision. 

Specifically, we reformulate the GAN inversion module to generate multiple hypotheses using different numbers of layers. Among the multiple hypotheses $\{y_1, ..., y_N\}$ with the number of hypotheses $N$, we decide the most plausible one based on FID scores, which enables the inversion results to be in natural data distributions. We define the latent code for $n$-th attempt as $z^n_I$, which can be found by minimizing the distance function between $\text{down}(y_n)$ and $r_{\mathcal{T} \rightarrow \mathcal{S}}$ as 

\begin{equation}
    z^n_{I} = \text{argmin}_{z} {\mathcal{D}(\text{down}(y_n), r_{\mathcal{T} \rightarrow \mathcal{S}};W^n_{I})},
\end{equation}

where $\text{down}(\cdot)$ is the downsampling operator, and $y_n=F(z;W^n_{I})$. $\mathbf{W}_{I}$ is an inversion network parameter, so $\mathbf{W}^n_{I}$ is the parameters of $n$-th attempt. $\mathcal{D}(\cdot,\cdot)$ is the distance function, which can be L1, L2, or perceptual loss function~\cite{johnson2016perceptual}. 
By measuring the FID scores of reconstructed images such as $k_n$, and finding the minimum, we eventually get the final translation image $g = y_{n^*}$, leading to $n^* = \text{min}_{n} {k_n}$. Figure~\ref{mgan_architecture} visualizes our aforementioned procedure.

\begin{figure*}[t]
\begin{center}
\includegraphics[width=1\linewidth]{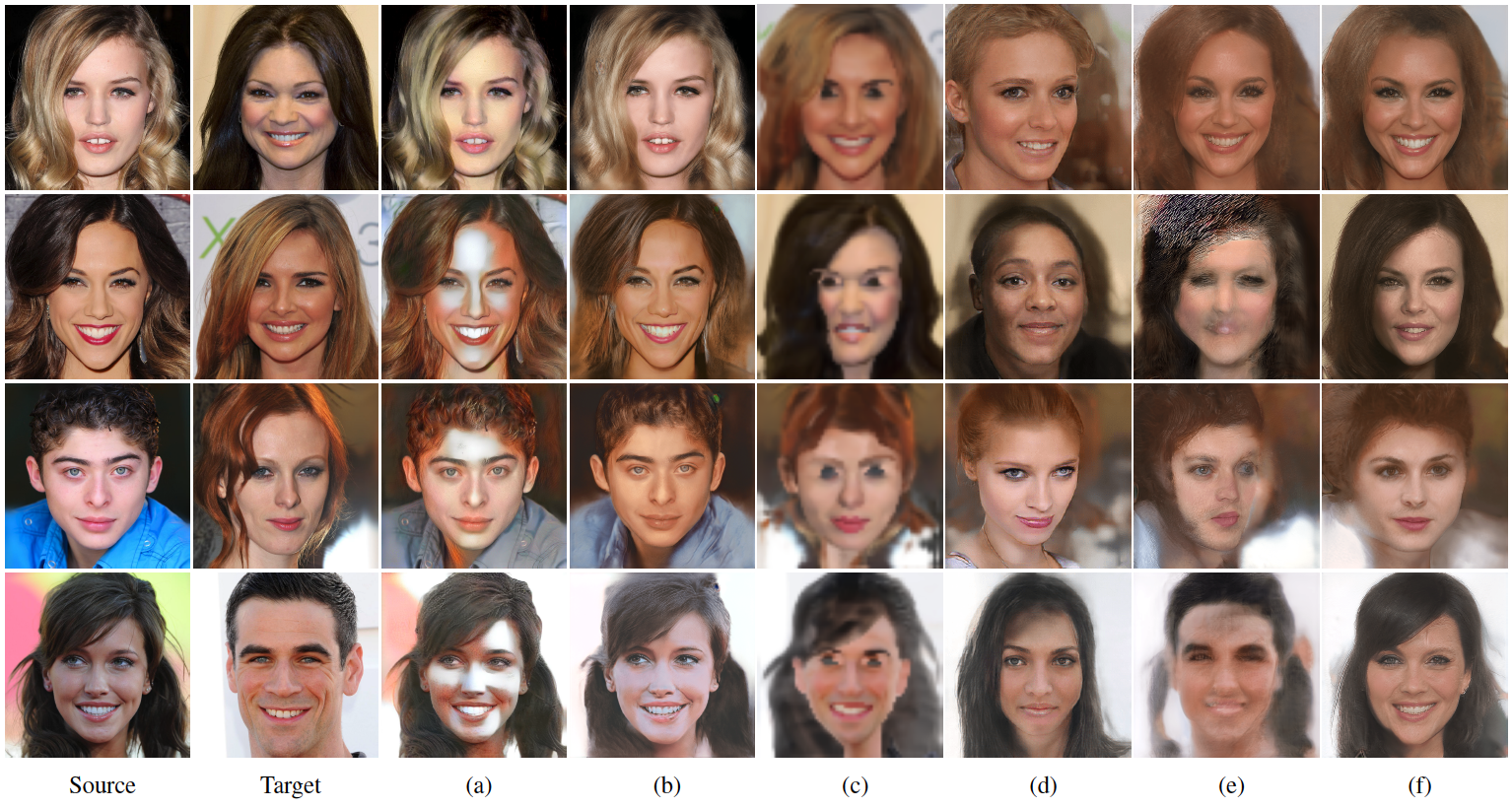}
\end{center}
\vspace{-15pt}
  \caption{\textbf{Comparison of our model with other existing exemplar-based image-to-image translation and style transfer methods on CelebA-HQ dataset.} Given source and target images, translation results by (a) Style Transfer~\cite{gatys2015neural}, (b) Image2StyleGAN~\cite{abdal2019image2stylegan}, (c) exemplar-warp, (d) CoCosNet-final~\cite{zhang2020cross}, (e) OEFT-final~\cite{kang2020online}, (f) Ours-fin. This shows that our networks can translate local features as well as global features from both structure and style. Note that (c), (d) are trained on CelebA-HQ mask2face dataset, including supervised segmentation mask on CelebA-HQ.} 
\label{celebahq_result}
\end{figure*}

\begin{figure}[t]
\begin{center}
\includegraphics[width=1\linewidth]{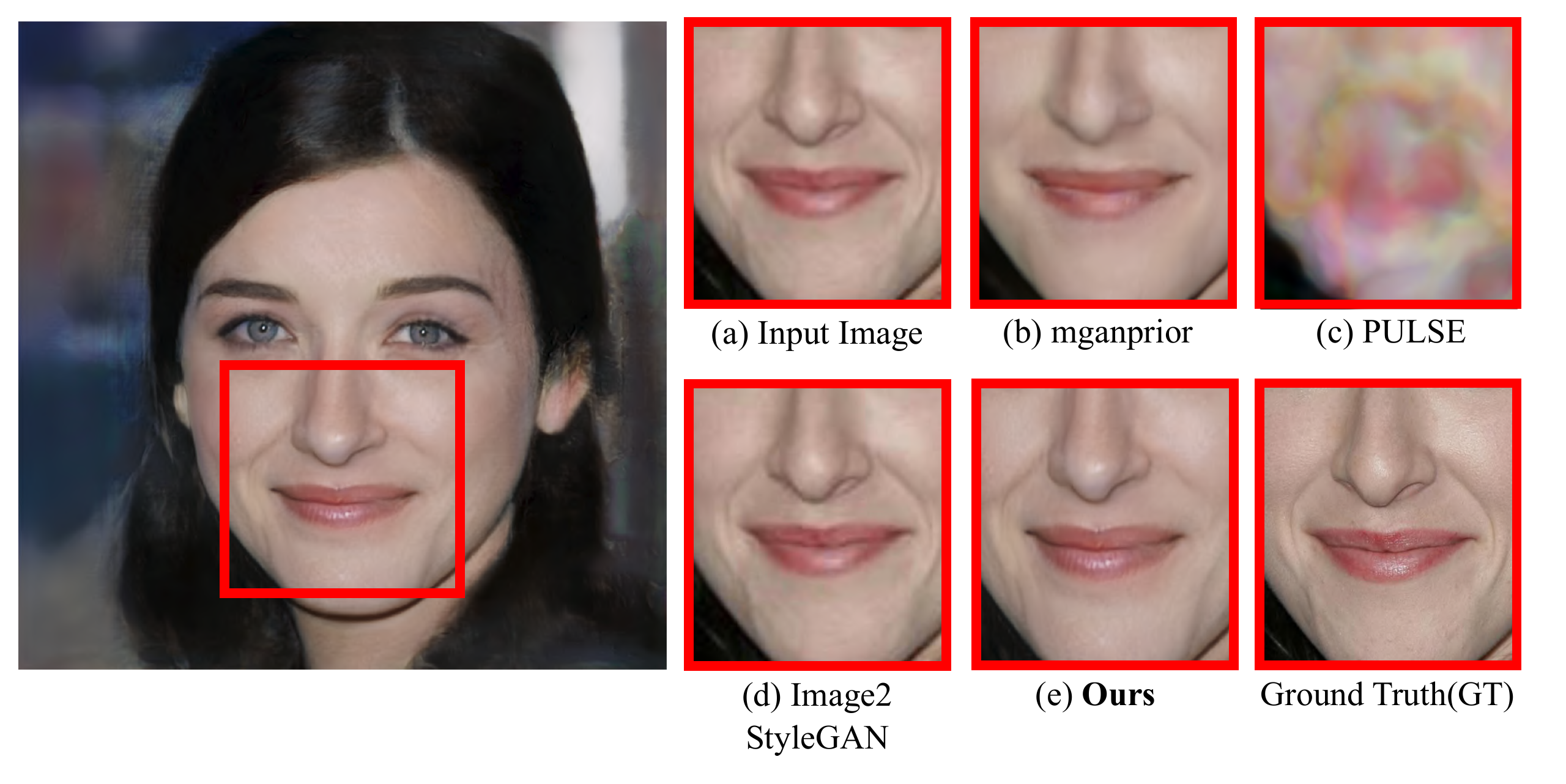}
\end{center}
\vspace{-15pt}
  \caption{\textbf{Comparison of GANs Inversion result with existing methods on CelebA-HQ dataset.} Our model generates competitive results with the smallest computational cost.} 
\label{gan_inversion_celebahq}
\end{figure}

\begin{figure}[t]
\begin{center}
\includegraphics[width=1\linewidth]{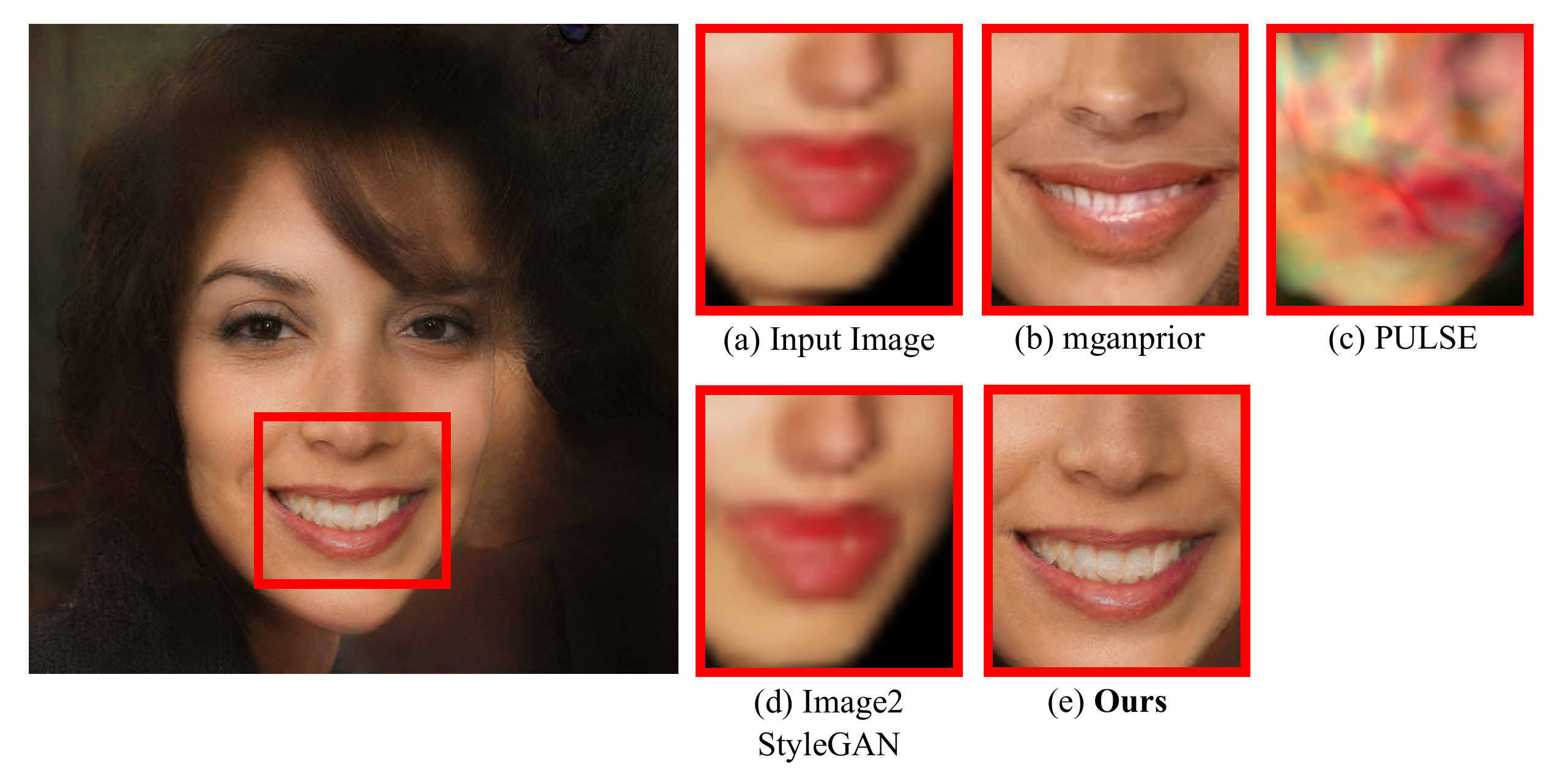}
\end{center}
\vspace{-15pt}
  \caption{\textbf{Comparison of GANs Inversion result with existing methods on warped image.} Our Multiple GAN Inversion result of warped image(generated by CoCosNet) super resolution, compared with existing methods. Our approach successfully generates the most realistic and clearest results.}
\label{gan_inversion_warped}
\end{figure}

\begin{figure*}[t]
\begin{center}
\includegraphics[width=1\linewidth]{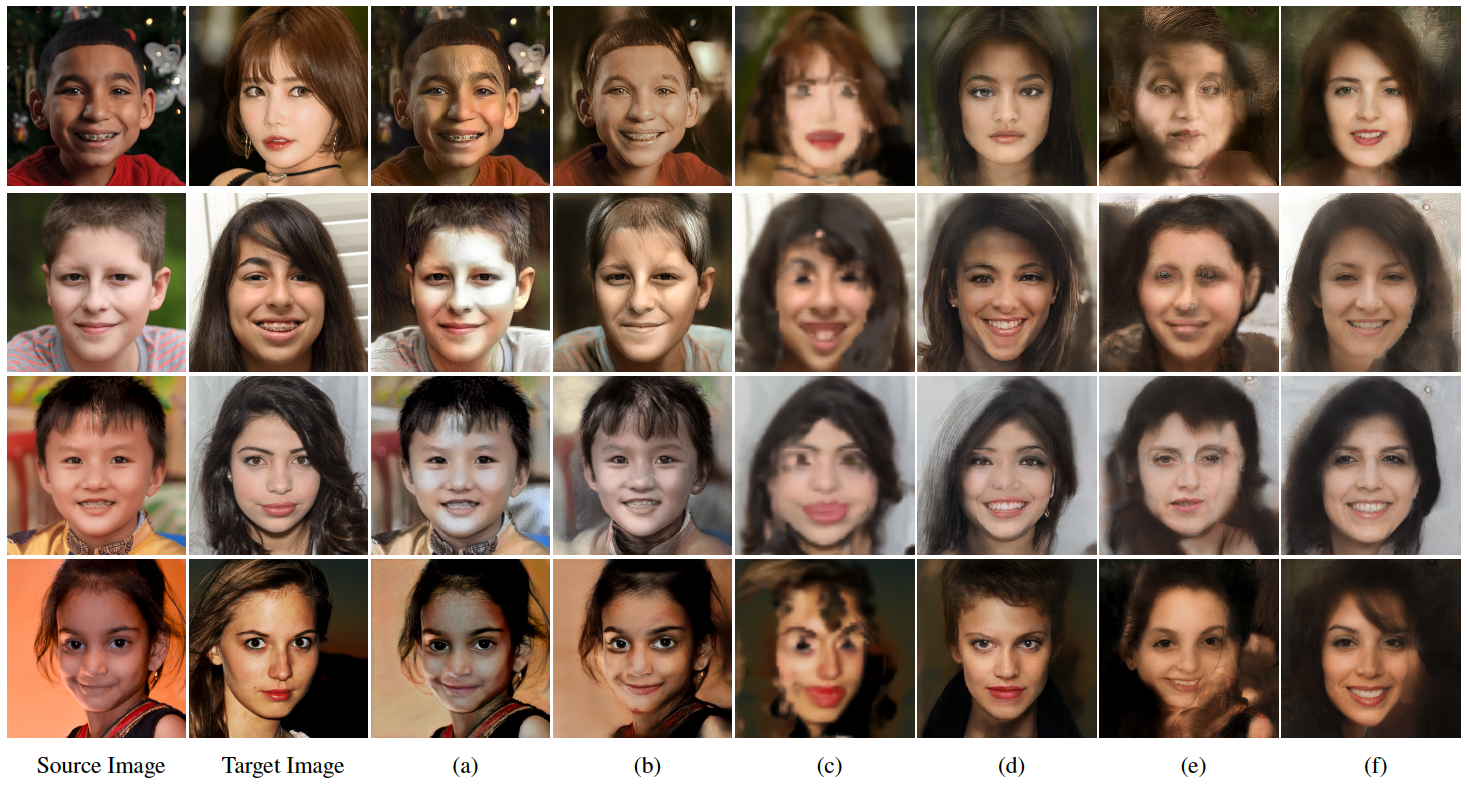}
\end{center}
\vspace{-15pt}
  \caption{\textbf{Comparison of our model with other existing exemplar-based image-to-image translation and style transfer methods on FFHQ dataset.} Given source and target images, translation results by (a) Style Transfer~\cite{gatys2015neural}, (b) Image2StyleGAN~\cite{abdal2019image2stylegan}, (c) exemplar-warp, (d) CoCosNet-final~\cite{zhang2020cross}, (e) OEFT-final~\cite{kang2020online}, (f) Ours-fin. This shows that our networks can translate local features as well as global features from both structure and style, and generates high-resolution images. Note that (c), (d) are trained on FFHQ dataset.} 
\label{ffhq_result}
\end{figure*}

\section{Experiments}
\subsection{Implementation Details}
In this section, we report implementation and training details to reproduce our experiments.

We first summarize implementation details of our approach, especially in cross-domain correspondence networks and multiple GANs inversion. In the correspondence module, we used the CoCosNet default setting for experiments. We used an Adam solver with $\beta_1 = 0$, $\beta_2 = 0.999$. Following the TTUR, we set imbalanced learning rates, $1e-4$ and $4e-4$, respectively.

\begin{table}
\begin{center}
\begin{tabular}{|l|c|c|}
\hline
Model & celebahq & FFHQ \\
\hline\hline
Style Transfer & 229.723 & 259.865 \\
Image2StyleGAN & 233.299 & 231.660 \\
CoCosNet(warp) & 383.441 & 377.943 \\
CoCosNet(result) & 220.359 & 235.772 \\
OEFT(result) & 248.193 & 272.415 \\
Ours(result) & \textbf{202.349} & \textbf{220.951} \\

\hline
\end{tabular}
\end{center}
\vspace{-5pt}
\caption{\textbf{Quantitative Evaluation of our model.} We used Fréchet Inception Distance(FID) to measure the performance of various network structures.}
\label{quantitative_overall}
\end{table}

\begin{table}
\begin{center}
\begin{tabular}{|l|c|c|}
\hline
Model & celebahq & warped \\
\hline\hline
mGANprior & 74.749 & 252.952 \\
PULSE & 449.530 & 470.946 \\
Image2StyleGAN & 71.114 & 363.223 \\
Ours & \textbf{50.427} & \textbf{179.522} \\
\hline
\end{tabular}
\end{center}
\vspace{-5pt}
\caption{\textbf{Quantitative Evaluation our GAN Inversion.} We used Fréchet Inception Distance(FID) to measure the performance of various network structures.}
\label{quantitative_mgan}
\end{table}

For the Multiple GAN Inversion module, we basically try to follow the default setting of mGANprior with some  modifications. We hypothesize composing layers ranging from 4 to 8, Number of the latent codes ranging from 10 to 40. The up-sampling factor is 4 for processing 256 $\times$ 256 to 1024 $\times$ 1024 images. We used PGGAN-Multi-Z and StyleGAN loss in mGANprior. For the distance function, we used L2 and perceptual Loss, as in mGANprior. We used a batch size of 1.

\subsection{Experimental Setup}
\textbf{Datasets.} We used two kinds of datasets to evaluate our method, namely CelebA-HQ\cite{liu2015deep} and Flickr Faces HQ(FFHQ)\cite{karras2019style}. During the optimization, input images were resized into 256 $\times$ 256. 

\textbf{Baselines.} 
We compared our method with recent state-of-the-art exemplar-based image-to-image translation methods such as CoCosNet~\cite{zhang2020cross}, OEFT~\cite{kang2020online}, Image2StyleGAN~\cite{abdal2019image2stylegan}, and Style Transfer\cite{gatys2015neural}. In addition, we also evaluate our GAN inversion module in comparison with mGANprior~\cite{gu2020image}, PULSE~\cite{menon2020pulse}, and Image2StyleGAN~\cite{abdal2019image2stylegan}, which have been state-of-the-art in GAN inversion. 

\begin{figure}[t]
\begin{center}
\includegraphics[width=1\linewidth]{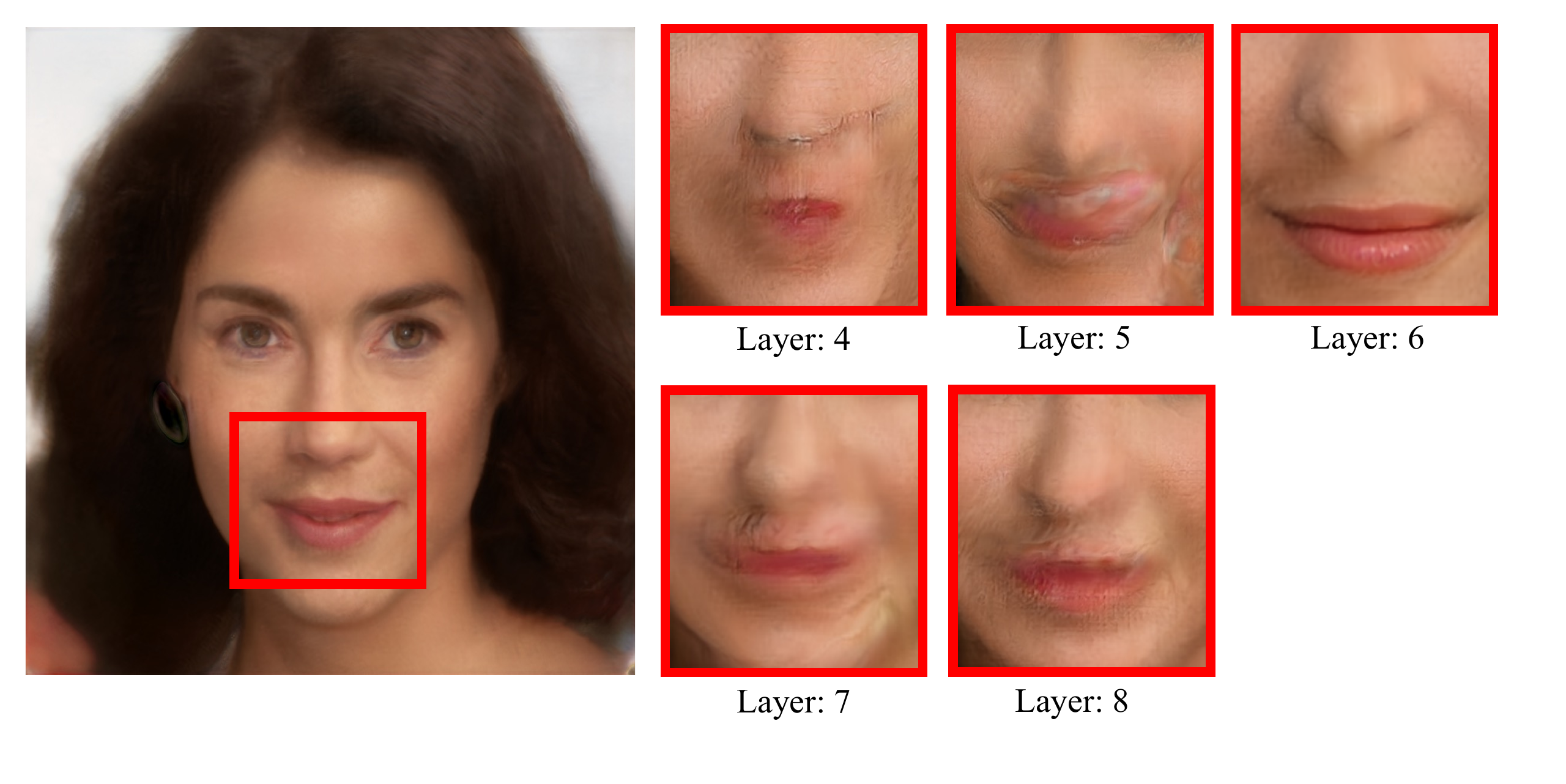}
\end{center}
\vspace{-18pt}
  \caption{\textbf{Ablation Study.} Ablation study result. Details can be found at the Experimental Results section.}
\label{ablation_study}
\end{figure}

\subsection{Experimental Results}
Since our framework consists of two sub-networks, Cross-domain Correspondence Networks and Multiple GAN Inversion, in this section, we evaluate the two modules separately, specifically \emph{exemplar-warp} denotes the results by only Cross-domain Correspondence Networks module~\cite{zhang2020cross}, and \emph{ours-fin} denotes the inversion results by using Multiple GAN Inversion module. We evaluate our final results as \emph{ours-fin}. Figure~\ref{celebahq_result} and Figure~\ref{ffhq_result} shows the experimental results.

\begin{figure}[t]
\begin{center}
\includegraphics[width=0.7\linewidth]{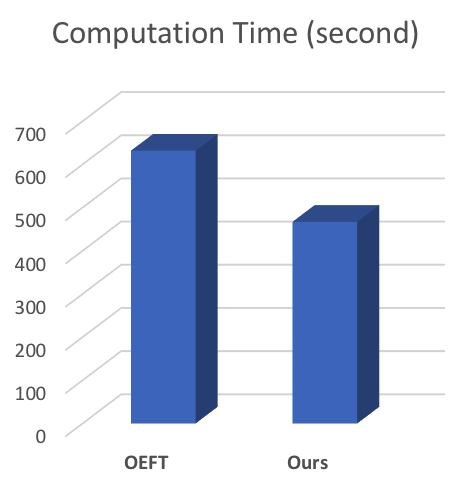}
\end{center}\vspace{-15pt}
  \caption{Computation time result.}
\label{computation}
\end{figure}

\textbf{Qualitative evaluation.} We evaluate our MGI module in Figure~\ref{gan_inversion_celebahq} and Figure~\ref{gan_inversion_warped}, where we show that our generated results are more realistic and plausible than state-of-the-art models. For instance, our method better generates features including the eyes or wrinkles as well as the overall structure, clearly outperforming the state-of-the-art methods. 
In addition, our approach has shown the high generalization ability to any unseen input images, which the other previous methods often fail.
Table~\ref{quantitative_overall} and Table~\ref{quantitative_mgan} show quantitative evaluation result, which used Fréchet Inception Distance (FID) to measure the performance, between original dataset (CelebA-HQ, FFHQ) and generated image distribution. In Table~\ref{quantitative_mgan}, we compared the original dataset(CelebA-HQ, warped image(CelebA-HQ original image 1024X1024)) and generated image(super-resolution) distribution.

\textbf{Ablation Study.} In this section, experiments are conducted to evaluate the effectiveness of Multiple GAN Inversion(MGI) networks of exemplar-based image-to-image translation. We evaluate the impact of each composing layers, ranging from 4 to 8. Figure~\ref{ablation_study} shows the ablation study result. We used PGGAN-Multi-Z loss for these experiments, and we fixed the number of latent codes to 30.

\textbf{Computation Time} We evaluate our MGI module's computation time with state-of-the-art model, OEFT~\cite{kang2020online} and CoCosNet~\cite{zhang2020cross}. CoCosNet reported in their paper using 8 32GB Tesla V100 GPUs, and it takes roughly 4 days to train 100 epochs on the ADE20k dataset. However, OEFT and MGI uses online optimization and pre-trained gan model, respectively. Compared to OEFT with using CelebA-HQ dataset, OEFT takes 632 seconds, and our model takes 467.09 seconds. Figure~\ref{computation} shows our evaluation result.

\begin{figure}[t]
\begin{center}
\includegraphics[width=0.55\linewidth]{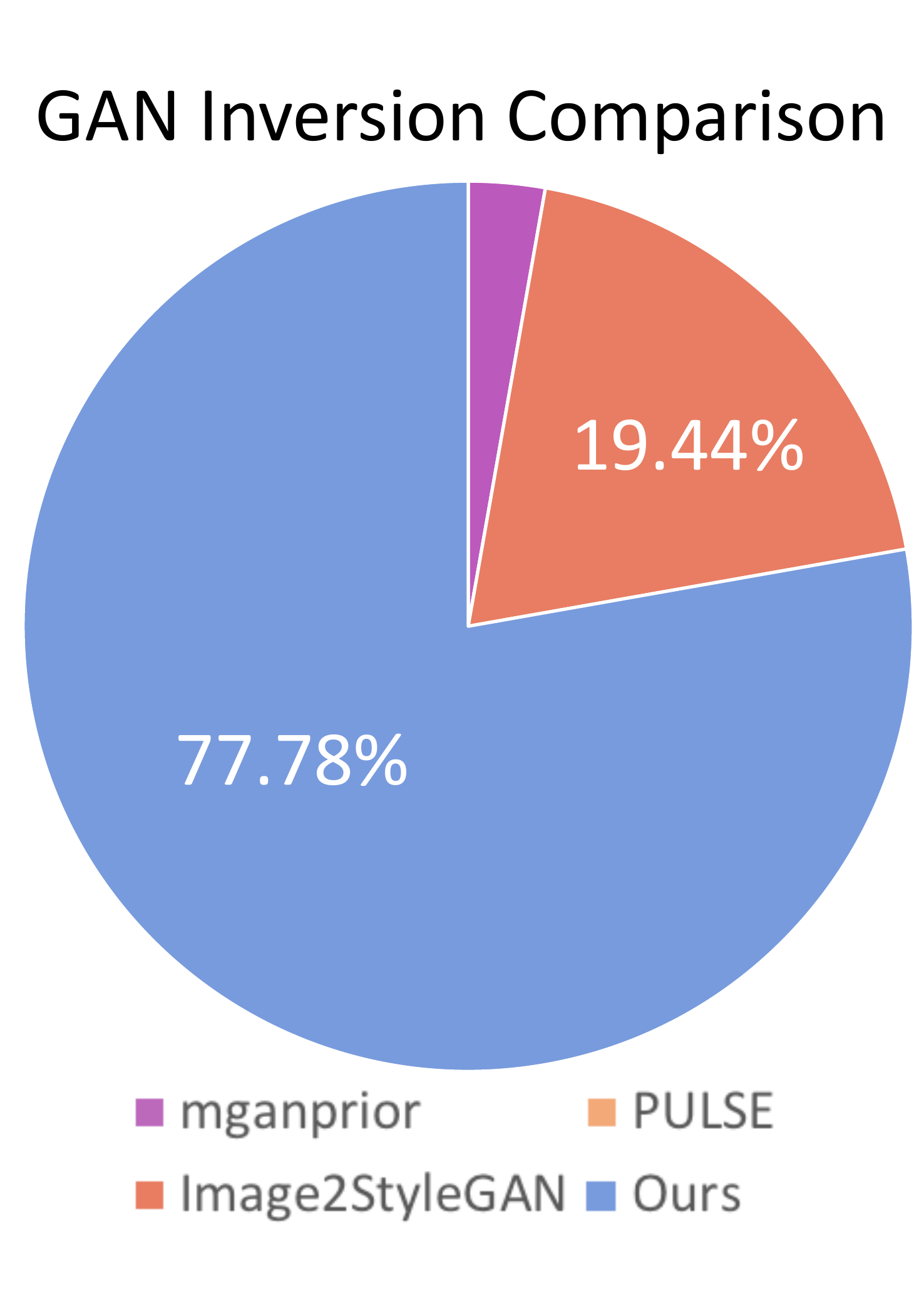}
\end{center}\vspace{-15pt}
  \caption{User Study result.}
\label{fig_userstudy}
\end{figure}

\textbf{User study.} We also conducted a user study on 80 participants. Figure~\ref{fig_userstudy} shows our user-study result. The source, target, and warped images are conducted randomly in 15 sets. The users are asked to vote on the 1024 $\times$ 1024 super-resolution outputs of our model, and outputs of other GAN Inversion super-resolution models according to the following question: which image quality is better; which image preserves the details of the source image and target image? In OEFT~\cite{kang2020online}, there are 63.27\% users that prefer the image quality produced from our method. In our MGI model, 77.78\% of users prefer our image quality, and most respondents prefer the structure of the source image and style of the target image.

\section{Conclusion}
In this paper, we proposed a novel framework, Multiple GAN Inversion(MGI) for Exemplar-based Image-to-Image Translation. Our model generates a more natural and plausible image through the proposed Multiple GAN Inversion, using a self-deciding algorithm in choosing the hyper-parameters for GANs inversion. We formulate the overall network from the two sub-networks, Cross-domain Correspondence Networks and Multiple GAN Inversion. Our approach applies high generalization ability to unseen image domains, which is one of the major bottlenecks of state-of-the-art methods. Experimental results showed superiority in the proposed method compared to existing state-of-the-art exemplar-based image-to-image translation methods.

{\small
\bibliographystyle{ieee_fullname}
\bibliography{egbib}
}

\end{document}